\documentclass[conference]{IEEEtran}

\usepackage[numbers,sort&compress]{natbib}

\ifCLASSINFOpdf
\usepackage[pdftex]{graphicx}
\else
\fi

\usepackage{amsmath}

\usepackage{subfigure}
\usepackage{multirow}
\usepackage{algorithm}
\usepackage{algorithmic}

\begin{document}

\title{Dynamic Island Model based on Spectral Clustering in Genetic Algorithm}



\author{\IEEEauthorblockN{Qinxue Meng\IEEEauthorrefmark{1},
		Jia Wu\IEEEauthorrefmark{1},
		John Ellis\IEEEauthorrefmark{2} and
		Paul J. Kennedy\IEEEauthorrefmark{1}}
	\IEEEauthorblockA{\IEEEauthorrefmark{1}
		Centre for Artificial Intelligence, Faculty of Engineering and Information Technology, \\ University of Technology Sydney, Sydney, Australia \\
	}
	\IEEEauthorblockA{\IEEEauthorrefmark{2}
		School of Life Sciences, Faculty of Science, University of Technology Sydney, Sydney, Australia \\}
	\IEEEauthorblockA{\IEEEauthorrefmark{0} Email: \{Qinxue.Meng, Jia.Wu, John.Ellis, Paul.Kennedy\}@uts.edu.au }
}

\maketitle

\begin{abstract}
How to maintain relative high diversity is important to avoid premature convergence in population-based optimization methods. Island model is widely considered as a major approach to achieve this because of its flexibility and high efficiency. The model maintains a group of sub-populations on different islands and allows sub-populations to interact with each other via predefined migration policies. However, current island model has some drawbacks. One is that after a certain number of generations, different islands may retain quite similar, converged sub-populations thereby losing diversity and decreasing efficiency. Another drawback is that determining the number of islands to maintain is also very challenging. Meanwhile initializing many sub-populations increases the randomness of island model. To address these issues, we proposed a dynamic island model~(DIM-SP) which can force each island to maintain different sub-populations, control the number of islands dynamically and starts with one sub-population. The proposed island model outperforms the other three state-of-the-art island models in three baseline optimization problems including job shop scheduler problem, travelling salesmen problem and quadratic multiple knapsack problem.

\end{abstract}

\IEEEpeerreviewmaketitle

\section{Introduction}
Optimization refers to finding best available solutions~($R_{k}$) for a given problem~($Q$) in a possible solution space~($R_{n}$) where~$n>>k$ within reasonable time. Effectively and efficiently optimizing a given problem is of much importance in fine-tuning parameters in machine learning and searching for potentially global solutions for NP-hard problems~\cite{gendreau2010handbook} which cannot be solved within polynomial time.

Population-based evolutionary optimizations, initially inspired by biological process, are considered as a major way of solving complex optimization problems. In the past few decades, a number of population-based evolutionary algorithms are proposed such as simulated annealing~(SA), differential evolution~(DE), Artificial Immune System(AIS) \cite{Wu:IJCNN131,Wu:IJCNN132}, particle swarm optimization~(PSO), bee colony optimization~(BCO), genetic algorithms~(GAs), population-based increamental learning~(PBIL) and ant colony optimization~(ACO)~\cite{lin2016simple}. Among them, SA, DE, PSO and BCO are roughly proposed to find solutions in continuous domains mainly for function optimization while GA, PBIL and ACO are mainly used for solving combinatorial optimization problems~\cite{vcrepinvsek2013exploration}. These methods optimize a given problem by maintaining a population of solutions from which the new solutions are created by using different types of evolution operators. For example, genetic algorithms create a new solution in the population of current generation from the mutation of a solution or a crossover of different parts of several solutions from the population of the previous generation.

A primary concern of population-based evolutionary algorithms is how to balance exploration and exploitation during the process of evolutionarily generating/searching new solutions~\cite{eiben1998evolutionary}. Given a solution space, exploration mainly aims to search and evaluate solutions in new regions while exploitation mainly aims to search and evaluate neighbours of previously evaluated solutions. A successful population-based evolutionary algorithm needs to establish a reasonable ratio between exploration and exploitation because exploration mainly works in the global area attempting to cover more regions and exploitation mainly works in a local area attempting to find best solutions in the local area. A good ratio of exploration and exploitation shows that an identified solution is the best in a sufficiently large area which means this solution approaches the globally optimized solution. How to control the ratio of exploration and exploitation is challenging because it is often implicit in population-based evolutionary algorithms and hard to control directly~\cite{vcrepinvsek2013exploration}. As a result, diversity~\cite{eiben1999parameter, back1996evolutionary, eiben1998evolutionary} is proposed and widely accepted within the evolutionary algorithms community as a measure for exploration and exploitation. This is because exploration is emphasized if the diversity of populations is high and exploitation is emphasized if the diversity of populations is low.

A great number of methods are proposed to balance exploration and exploitation by maintaining population diversity. The initial work of maintaining diversity is based on population control directly including maintaining a large population~\cite{mauldin1984maintaining}, reinitialize populations~\cite{li2011chaotic}, randomly introducing new individuals~\cite{wongseree2007thalassaemia} or replacing old individuals with new ones~\cite{jia2011effective}. However, it is hard to maintain diversity by doing operations on populations directly as in most cases, these methods treat the evolutionary process as a black box so as to be hard to guide the direction of evolution. Another way of maintaining diversity is through genetic operators such as selection, mutation, crossover, and other new proposed genetic operators or combining them together~\cite{chen2009preserving, mallipeddi2011differential}. Through these methods better control of the whole evolutionary process is possible. The initial fixed and constant rate of genetic operators are hard to determine their values in an adaptive way as evolution proceeds~\cite{zamuda2015self}. These methods often come with high computational complexity and may be trapped in local dilemma. Currently, island model~\cite{leitao2015island} is widely accepted as one of the most successful ways of maintaining diversity. This method is proposed to maintain sub-population in each island and individuals from different islands may migrate from one to the other. The benefits of island model are many. Firstly it is flexible. Islands can have different strategies to maintain diversity individually. Secondly, it is efficient. It supports parallel computation and can be integrated with MapReduce framework~\cite{salza2016elephant56} because islands are evolved independently. However, the major drawback of this method is that after a certain number of generations, different islands may maintain quite similar, converged sub-populations. This, on one hand, may contribute to the loss of diversity to some extent. Additionally, maintaining islands with similar, converged sub-populations is a waste of computational resources. Further, it is also challenging to determine how many islands should be maintained.

To deal with these issues, we propose a novel dynamic island model based on spectral clustering~(DIM-SP). The main principle of the proposed method is to introduce a new migration policy. Instead of having individuals migrate from one island to the other, the proposed migration policy is to allow all sub-populations to migrate together and then individuals are assigned to different islands by spectral clustering. The number of islands are determined by the number of clusters. The contributions of the proposed DIM-SP model can be summarized as
\begin{enumerate}
	\item The involvement of clustering enables DIM-SP model to explicitly maintain exploration among islands and focus on exploitation in islands. This provides an intuitive way for users to control the process of evolution;
	\item the proposed DIM-SP island model can avoid having islands maintain similar, converged sub-populations;
	\item the number of islands is dynamic and can be controlled by the number of clusters;
	\item DIM-SP can allocate more survival space for diversities with few individuals;
	\item DIM-SP can be initialized with one population in one island which simplifies the initialization of island model;
\end{enumerate}

To evaluate the proposed DIM-SP island model, we compared the performance of this model with three other topological island models including fully-connected, star-shape and ring island models on a set of benchmark optimization problems.

\section{Literature review}
\begin{figure}
	\centering
	\includegraphics[width=18pc]{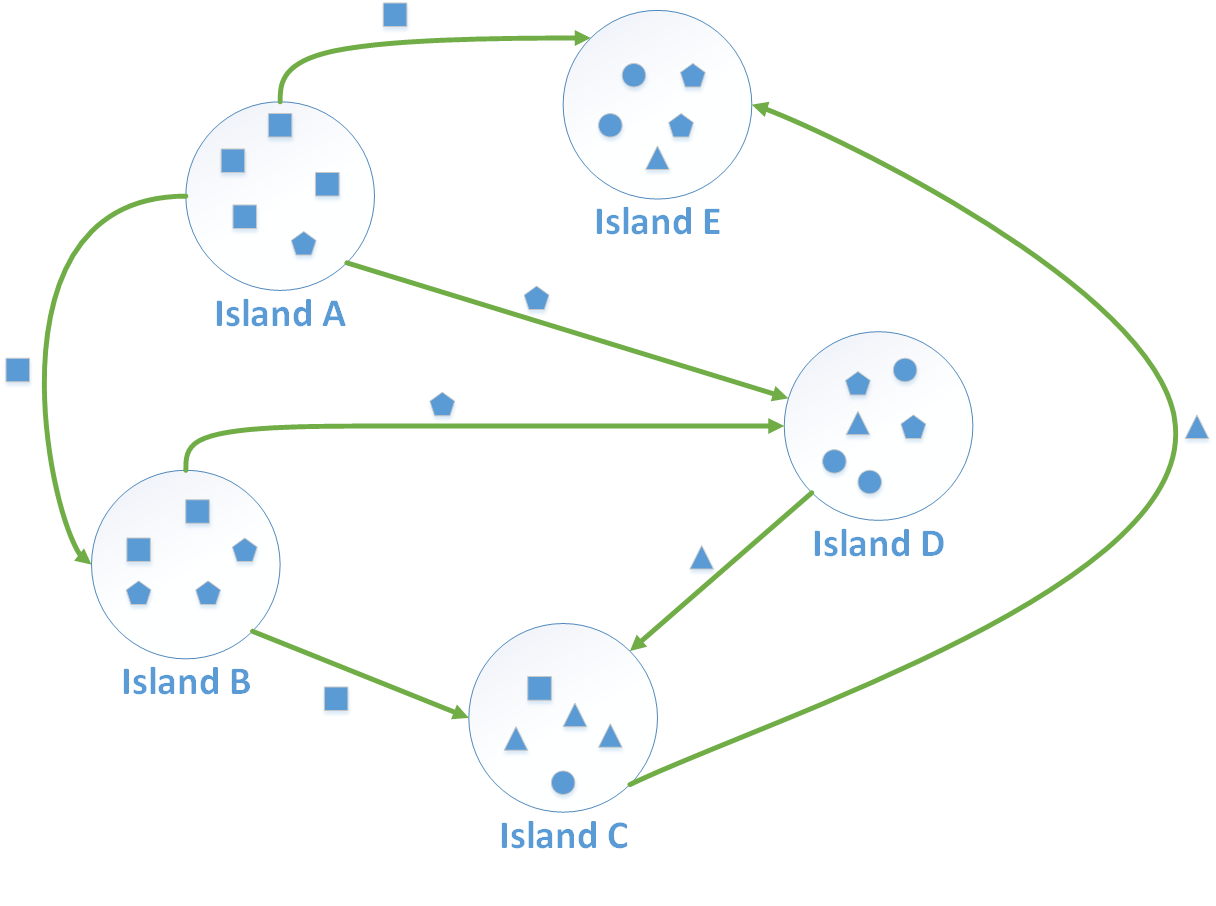}
	\caption{An example of traditional island model. There are five islands from A to E and each island maintains one sub-population. Individuals are allowed to migrate from one island to the other according to a specific migration policy. }
	\label{fig1}
\vspace{-0.2cm}
\end{figure}

The initial motivation of island model~(Fig.~\ref{fig1}) in genetic algorithm was derived from the considerations of natural evolution and parallel computing~\cite{martin1997island}. In 1964, Wright~\cite{wright1964stochastic} proposed a basic island model which maintains one sub-population in one island which is implemented in one cluster so as to achieve parallel evolution. In this model, each cluster carries out its own evolutionary algorithm, with frequent interchange of information among them via migration of individuals from one cluster to the others. The major success of island-model is to prevent inbreeding effectively so as to main relatively high diversity in the population~\cite{collins1991selection}. Inbreeding, referring to mating among similar individuals, cannot increase diversity and ultimately the diversity of the population decreases due to selection. To avoid this, island model isolates populations from each other leading them to explore different portions of the search space which has frequently proven to improve the results~\cite{tomassini2005spatially, alba2002heterogeneous}. Isolation among sub-populations is the key behind the island model so as to maintain diversity and this is determined by migration policies. Consequently, migration policies have frequently been the subject of study. The island model for parallel evolutionary algorithms were later converted to work sequentially~\cite{collins1991selection, tomassini2005spatially, alba2009cellular}. This removes the constraints of island model from hardware which provides flexible and dynamic control of island number.

Migration policies have included the number of migrant individuals, the frequency of the migration, the policy for selecting the individuals sent to other nodes and to be replaced by the received ones, the network topology or the synchronous and asynchronous nature of the communications~\cite{limmer2016comparison}. The early migration policies focused on exchange strategies amongst sub-populations including migrants replacing less fit individuals, randomly chosen individuals or the most similar individuals~\cite{alba1999survey}. These studies are based on a broad assumption that different islands can maintain different sub-populations. Many novel variants and extensions of migration policy were proposed which consider many different natural processes such as biological, physical and sociological processes~\cite{6718037, boussaid2013survey}. However, as evolution proceeds, it is often seen that different islands maintain quite similar sub-populations because as highly fit individuals migrate to sub-populations with relatively low fitness values, they are more likely to survive and breed which leads the whole population to converge to those new migrated individuals.

To deal with this issue, Ursem~\cite{ursem2000multinational} proposed a multinational genetic algorithm~(MGA) where sub-populations are self-formed on a hill-valley detection algorithm. Specially, solution space is considered as a landscape where sub-populations are formed and can move on. The direction of movement is determined by the best fitted individuals who are marked as government. Individuals can move across sub-populations and moving to near sub-populations gives high priority. MGA is successful in maintaining multiple, different peaks simultaneously as similar sub-populations move together and merged. However, there are still several weaknesses. The model is not that efficient. The moving speed of sub-populations in a given landscape is generally slow. If there are several similar peaks in one local area, sub-populations are likely to be trapped and move around these peaks. In addition, the movement direction is determined by individuals with the highest fitness within sub-populations. It is possible to see many sub-populations move towards to the same peak therefore losing diversity. The model was extended and improved by subsequent studies~\cite{baykasoglu2015constructive, ozsoydan2015multi, nasiri2016improved} which proposed more complex migration policies to maintain diversity. This practice decreased the model's efficiency further.

Another important contribution of MGA is to consider the distance between sub-populations and fitness landscape. Subsequently more complex island models were proposed~\cite{7368156, li2015history} including ring model, centralized and compete graph models. They consider the structure of islands as a network and proposed migration policy based on island topology. However, complex policies make the control of exploration and exploitation more difficult and implicit.

\begin{figure}
	\centering
	\includegraphics[width=18pc]{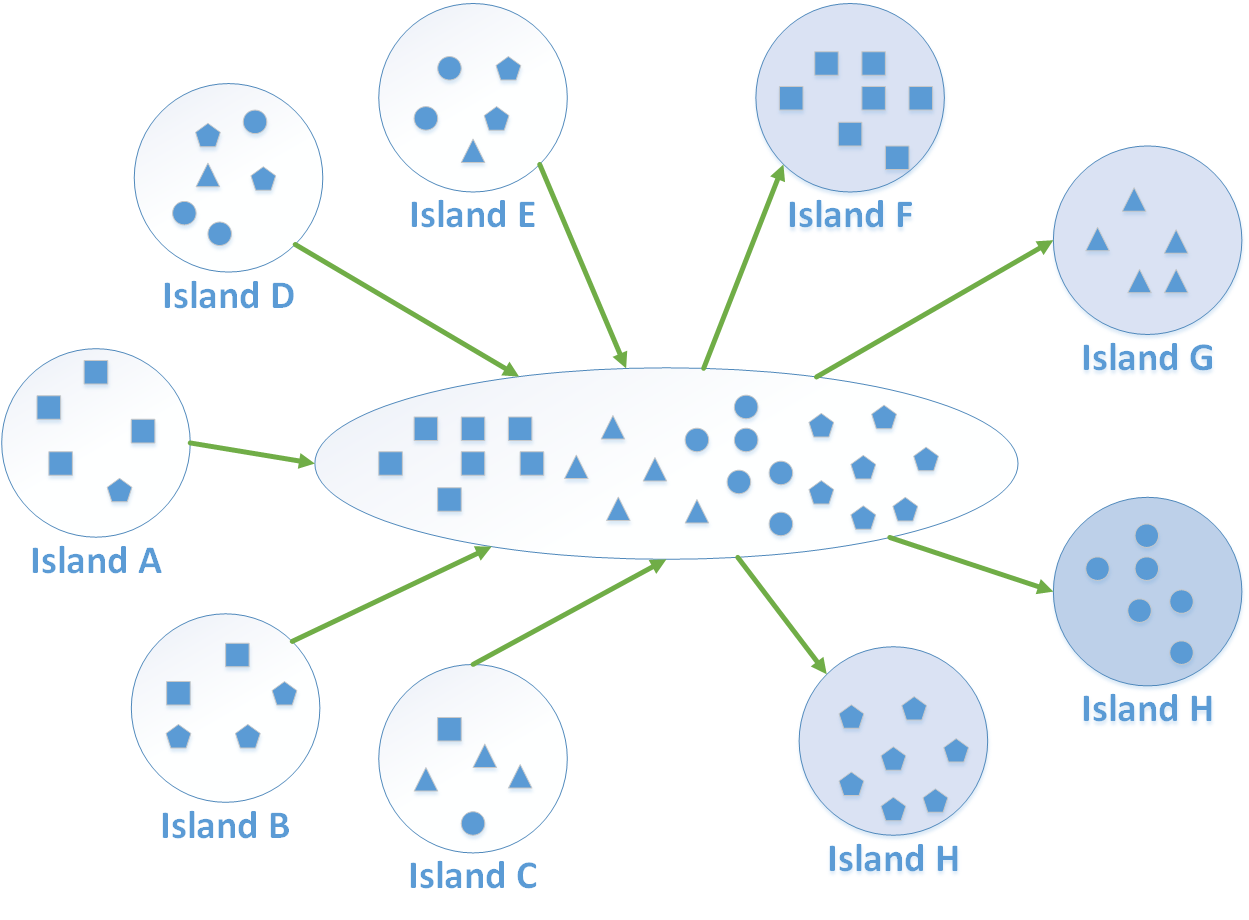}
	\caption{An example of the proposed dynamic island model based on spectral clustering~(DIM-SP). Periodically sub-populations from old Island A to E migrate together. Similar individuals are assigned to new Island F to H by the results of spectral clustering and old islands are destroyed. The number of new islands are determined by the number of clusters.}
	\label{fig2}
\vspace{-0.2cm}
\end{figure}

\begin{figure*}[!t]
	\centering
	\subfigure[Fully-connected island model]{\includegraphics[width=0.3\textwidth]{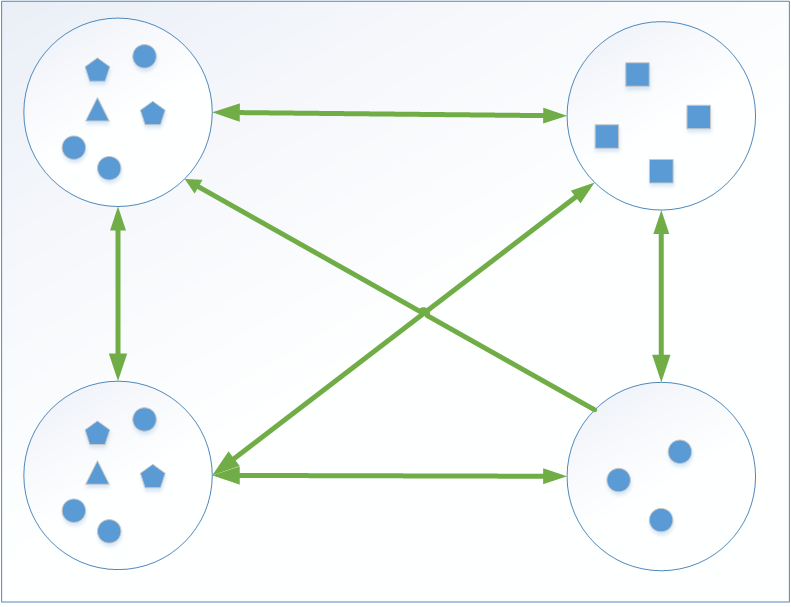}}
	\subfigure[Start-shaped island model]{\includegraphics[width=0.3\textwidth]{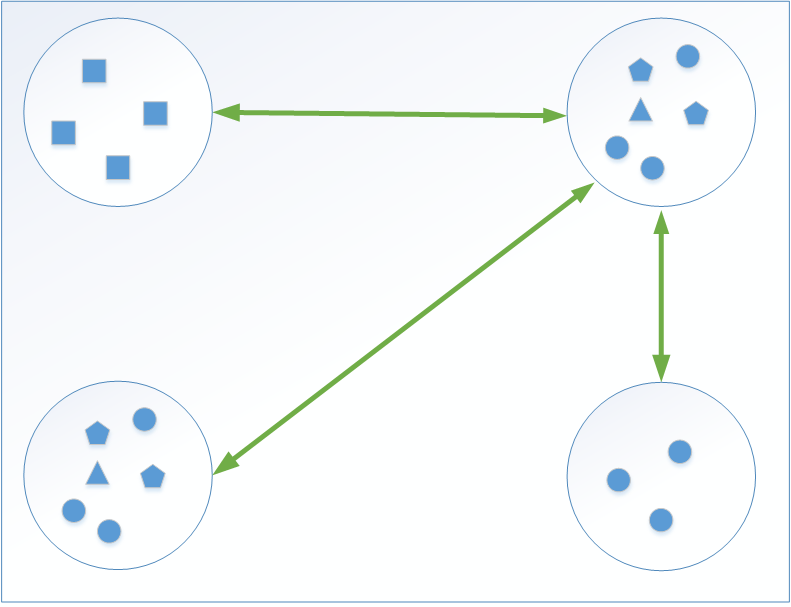}}
	\subfigure[Ring island model]{\includegraphics[width=0.3\textwidth]{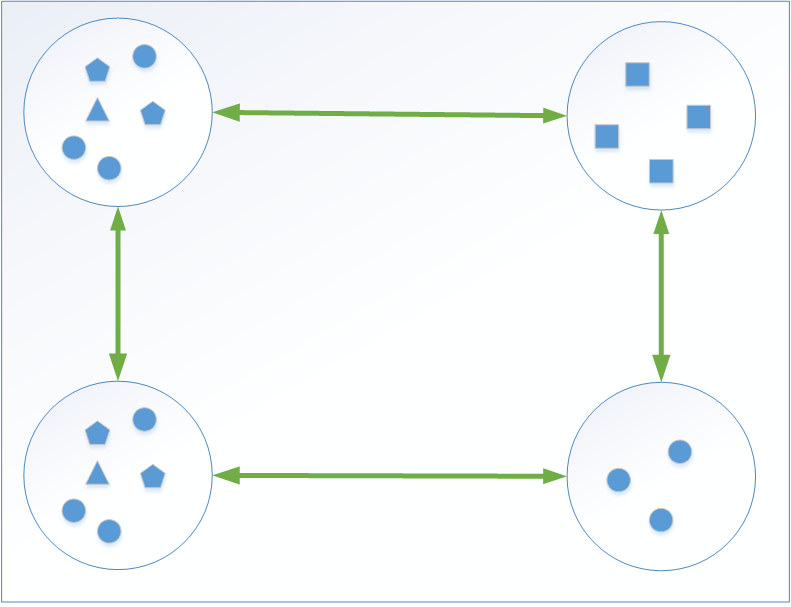}}
	\caption{Examples of the baseline, topological island models. (a) is a fully-connected island model where individuals can migrate from one island to the others; (b) is a star-shaped island model where there is a main island connecting all the other islands and individuals can migrate to the other islands via the main island; (c) is a ring island model where individuals can only migrate to islands next to them. }
	\label{fig5}
\vspace{-0.2cm}
\end{figure*}

As a result, we propose a dynamic island model based on spectral clustering~(DIM-SP) aiming to improve efficiency of island model, simplify migration policy and provide explicit control of exploration and exploitation~(Fig.~\ref{fig2}). In this study, the performance of the proposed DIM-SP island model is compared ot the other three topological island models including fully-connected, star-shape and ring island models~(Fig.~\ref{fig5}) on a set of benchmark optimization problems.

\section{Preliminaries}
This section begins with a discussion of the problem definition followed by a description of basic island model so as to emphasize the difference between traditional island model and the proposed dynamic island model based on spectral clustering~(DIM-SP).

\subsection{Optimization}
For a given problem $Q(f, S)$ where $S$ is a search space where elements represent potential solutions of the problem and $f$ is a single or combined fitness function to evaluate the performance of individual element $s^{*}$ where $s^{*} \in S$. The purpose of optimization is to find one or a set of most fitted elements $s^{*}$.

\subsection{Island model}
Consider an island model $\Phi(P, O, M)$, there are an overall population $P$~($P \subset S$) which is partitioned into $N$ sub-populations~$P_{1}, P_{2}, ..., P_{N}$ and each sub-population is assigned to one island. The sub-populations in different islands are evolved independently with the same/different evolving operators, marked by $O$. Sub-populations interact based on static/dynamic migration policy $M$.

\begin{algorithm}[b]
	\caption{Evolution process of basic island model}
	\label{alg:im}
	\begin{algorithmic}[1]
		\REQUIRE ~~\\
		Set an optimization problem $Q(f, S)$; \\
		Set $num\_island = n$;                 \\
		Set $P=\{P_{1},P_{2},\ldots,P_{n}\}$;  \\
		Set $O=\{O_{1},O_{2},\ldots,O_{n}\}$;  \\
		Set migration policy $M$;
		\ENSURE ~~\\
		\FOR {$i=1$ to $n$}
		\STATE $P_{i} = O_{i}(P_{i});$
		\FOR {$s \in P_{i}$}
		\STATE $v_{s} = f(s);$
		\FOR {$k=1$ to $n$}
		\IF  {$M(s, v_{s}, P_{i}, P_{j}) == 1$}
		\STATE $P_{j}= P_{j} \cap {s}$;
		\STATE $P_{i}= P_{i} \setminus {s}$;
		\ENDIF
		\ENDFOR
		\ENDFOR
		\ENDFOR
		\STATE $s^{*} = s^{*} \cap {best(P_{i})}$;
	\end{algorithmic}
\end{algorithm}

\subsection{Evolutionary process}
The principle of island model is to create isolated sub-populations so as to maintain high diversity. A migration policy is applied to allow sub-populations to interact with others for finding potential global solutions from local solutions. The evolutionary process of island model can be described in Algorithm~\ref{alg:im}. For every generation, each sub-population, e.g.~$P_{i}$ evolves and is evaluated independently. The migration policy then checks whether elements in this sub-population needs to migrate and which population to go to. At the end of each generation, the elements of best fit will be kept in $s^{*}$. The algorithm behind the basic island model shows that migration policy adds to the complexity of the computation which inspires us to propose a more efficient one.

\begin{figure*}
	\centering
	\includegraphics[width=40pc]{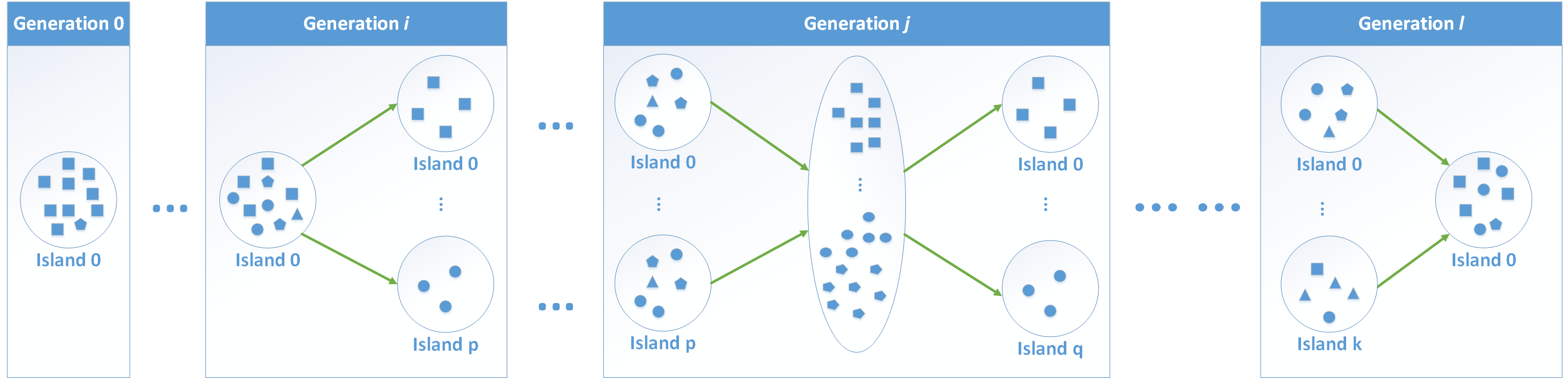}
	\caption{Evolutionary process of the proposed dynamic island model based on spectral clustering~(DIM-SP). It starts from one island with an initialized sub-population. As evolution proceeds to Generation $i$, sub-population of Island 0 migrate to new islands based on their similarities. As it continuous to Generation $j$, sub-populations migrate together and then migrate to different islands based on similarity. This centralization-clustering process is repeated until to the final Generation $l$ where all sub-populations migrate together and individuals with top highest fitness values are considered as solutions.}
	\label{fig3}
\end{figure*}

\section{Methodology of DIM-SP}
In contrast to the traditional island models that involve complex migration policies, the proposed island model simplifies it by introducing a centralization-clustering framework. This practice can sufficiently reduce the computational cost of island model. Meanwhile it can also provide simple, explicit control of exploration and exploitation so as to maintain diversity easily. The whole evolutionary process of the proposed dynamic island model based on spectral clustering~(DIM-SP) is illustrated in~Fig.~\ref{fig3}.

The proposed island model starts with one island instead of many which relieves the workload of initializing multiple islands in traditional island models. In DIM-SP model, Island $0$ is initialized in Generation $0$.

When the proposed model runs to a specific generation, say $i$, the Island $0$ is split into a set of new islands, labelled with $0, \ldots, p$ by spectral clustering based on the similarities of individuals. Similar individuals are assigned to the same island while dissimilar ones are assigned to different islands. The number of new islands are determined by the number of clusters which can be controlled dynamically. In this way, increasing the number of clusters or decreasing the pressure of clustering can promote exploration. After splitting, each new island evolves independently.

As it continues to Generation $j$, individuals of all islands~$0,\ldots,p$ migrate together and they are clustered based on their similarities into new islands~$0, \ldots, q$ where $p$ and $q$ can be equal or not. New islands are assigned a set of evolution operators randomly including selection, mutation, crossover and so on. Each centralization and clustering, like Generation $j$ is marked as one epoch. This migration policy has several benefits. Firstly, because each island has a limitation on sub-population size, individuals in large clusters are more likely to be removed and those in small clusters will be kept. Secondly, this policy increases the selection pressure on large clusters as they are mature communities and decrease the selection pressure on small clusters. In this way, individuals of minor clusters can have more survival space to grow so as to avoid die out before maturity. This can contribute to maintaining diversity to some extend. In large clusters, well fitted individuals can survive so as to promote exploitation.

After a number of epochs, in the final Generation $l$, all sub-populations migrate together and top ranked individuals are considered as potential global optimized solutions.

As a result, the pseudocode of the proposed DIM-SP model is described in Algorithm~\ref{alg:pim}. The model is initialized with one island and starts to evolve. If Generation $i$ is an epoch, the model starts to apply the centralization-clustering migration policy; otherwise, each island is evolved independently. In epochs, individuals from all islands $P_{1}, \ldots, P_{n}$ migrate together marked by $P_{a}$. Matrix $W$ is the similarity matrix of individuals calculated by similarity function $\varTheta$ and spectral clustering~($\varLambda$) is applied to cluster population~($P_{a}$) into a new set of sub-populations~($P$) based on the similarity matrix~($W$). Each island will be assigned a set of evolution operators by function~$\varUpsilon$ and the number of islands will be updated.

\begin{algorithm}[!htb]
	\caption{Evolution process of DIM-SP model}
	\label{alg:pim}
	\begin{algorithmic}[1]
		\REQUIRE ~~\\
		Set an optimization problem $Q(f, S)$; \\
		Set $n = 1$;                 \\
		Set $P=\{P_{1}\}$;  \\
		Set $O=\{O_{1}\}$;  \\
		Set migration policy $M$;
		\ENSURE ~~\\
		\FOR {$i=1$ to $n$}
		\IF {$i$ in $epochs$}
		\STATE $P_{a} = P_{1} \cup P_{2} ... \cup P_{n}$;
		\STATE $W = \varTheta(P_{a})$;
		\STATE $P = \varLambda(W)$;
		\STATE $O = \varUpsilon(P)$;
		\STATE $n = |P|$;
		\ELSE
		\STATE $P_{i} = O_{i}(P_{i});$
		\ENDIF
		\ENDFOR
		\STATE $s^{*} = s^{*} \cap {best(P_{i})}$;
	\end{algorithmic}
\end{algorithm}

\subsection{Similarity measure}
Another advantage of the proposed DIM-SP model controls individual migration policy based on the similarities of individuals which are used to evaluate population diversity. In other words, the migration policy of the proposed island model works on population diversity directly.

There are two major ways of measuring diversity of a given population based on individual similarities. Firstly, diversity refers to the difference amongst individuals. In this way, it is a measure of the similarities between individuals. If the genome of individuals are continuous values, the similarities of individuals can be measured by a distance measure such as Euclidian distance or cosine. If the genome of individuals are continuous values, the similarities of individuals can be measured based on their difference. For example, given two individuals $\textbf{x}$ and $\textbf{y}$ with genome length as $n$.
\begin{equation}
s_{f}(\textbf{x}, \textbf{y}) =
\begin{cases}
1, & \text{if}\ x_{i} = y_{i},  \\
0, & \text{otherwise},
\end{cases}
\end{equation}	
Diversity can also refer to the difference between fitness values of individuals and tries to retain best fitted individuals. The DIM-SP model can integrate different types of similarities to achieve different types of diversities.

In this study, we evaluate the proposed island model in combinational domain so we choose difference-based similarities of individuals.

\subsection{Spectral clustering}
Spectral clustering is introduced to perform clustering based on an individuals' similarity rather than other methods such as $k$-means, hierarchical clustering or DBSCAN because it can achieve a global optimization solution, is easily implemented and supports parallel computing~\cite{von2007tutorial}.

In this study, we use normalized spectral clustering to do clustering. Given a similarity matrix $W$ of individuals, the normalized Laplacian matrix $L$ is defined as
\begin{equation}
	L = I - D^{-1/2}WD^{-1/2}
\end{equation}
where matrix $I$ is the identity matrix and $D$ is the diagonal matrix of $W$. The first leading $k$ smallest eigenvalues and each eigenvalue is mapped to a cluster.

\subsection{Evaluation}
We compare the performance of DIM-SP with the other three state-of-the-art island models including ring, start-shaped and fully-connected island model with the same configuration of evolution operators and migration policy on a set of combinatorial optimization problems including job shop scheduler~(JSSP), Travelling Salesman~(TSP) and the quadratic multiple knapsack~(QMKP) problems. The performance of results were evaluated by fitness score and population diversity. The population diversity, in this paper, is calculated via the average distance between individuals of best fit and the other individuals. The proposed method and the other compared methods runs for ten times with different random seeds. The results are the averaged results of ten-time running.

\begin{figure*}[!t]
	\centering
	\subfigure[Avg scores of JSSP Problem]{\includegraphics[width=0.3\textwidth]{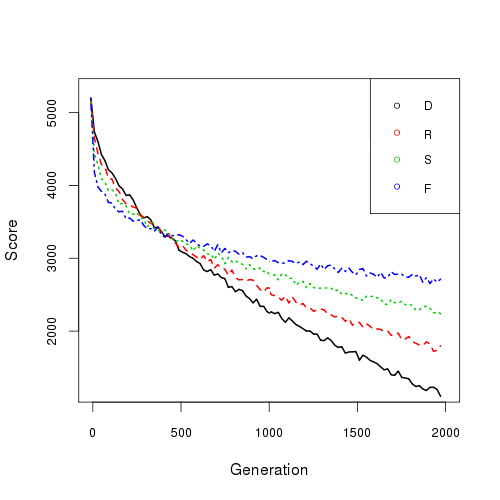}}
	\subfigure[Avg scores of TSP problem ]{\includegraphics[width=0.3\textwidth]{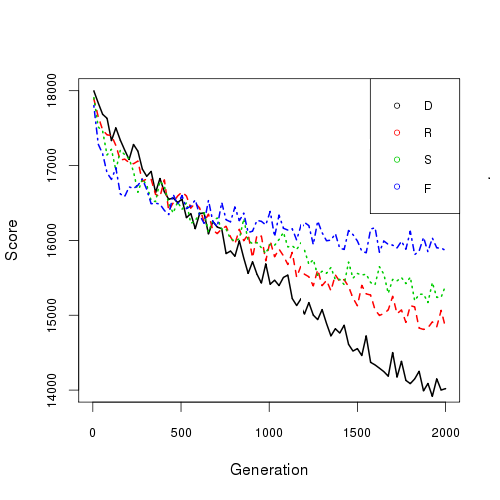}}
	\subfigure[Avg scores of QMKP problem ]{\includegraphics[width=0.3\textwidth]{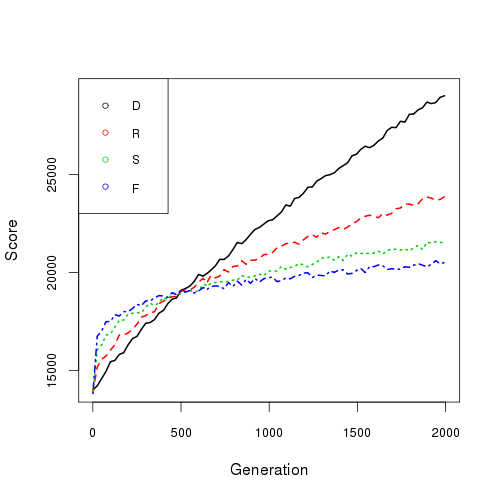}}
	\subfigure[Diversity of JSSP problem ]{\includegraphics[width=0.3\textwidth]{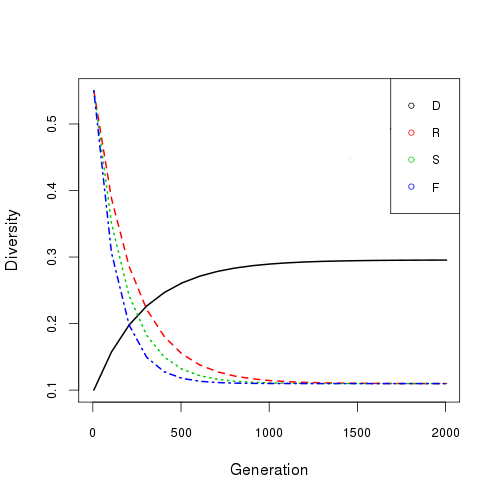}}
	\subfigure[Diversity of TSP problem ]{\includegraphics[width=0.3\textwidth]{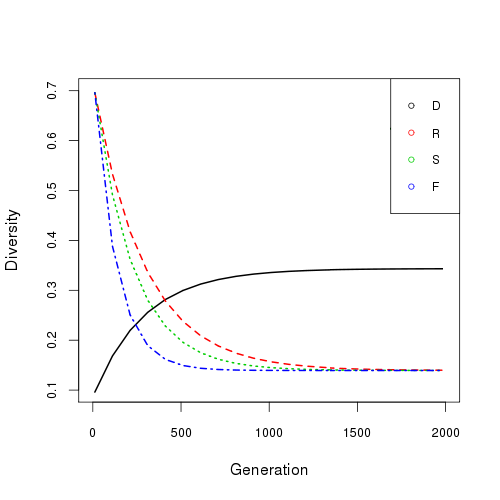}}
	\subfigure[Diversity of QMKP problem ]{\includegraphics[width=0.3\textwidth]{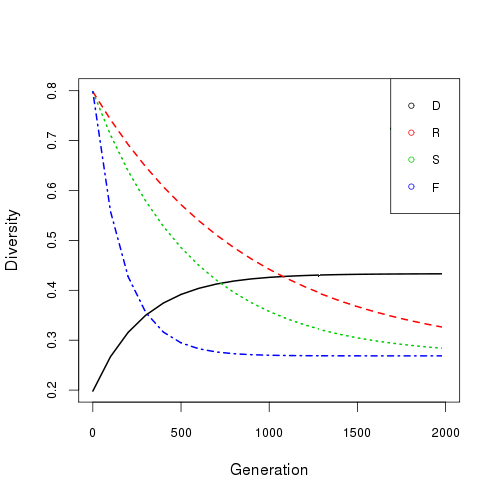}}
	\caption{Experimental results using DIM-SP island model. Specifically, (a), (b) and (c) illustrate the changes of average scores in different generations
		while (d), (e) and (f) illustrates the changes of diversities in the benchmark problems. In each sub figure, D, DIM-SP island model, F, Full-connected island model, S, Star-shape island model and R, Ring island model. The results of different models are marked with different colours and types of line.}
	\label{fig6}
\vspace{-0.2cm}
\end{figure*}

\section{Test functions}
The classic JSSP with the objective of makespan minimization consists of a set of $n$ jobs that needs to be processed on a set of $m$ machines. The processing of job $J_{j}$ on machine $M_{r}$ is called the operation $O_{jr}$. The solutions of this problem should satisfy two constraints: job should be processed on each machine according to a predefined sequence and each machine can process only one job at a time. The time needed for completion times for all operations is called makespan, denoted as $C$ and the objective of the problem is to minimize $C$. In this paper, we evaluate the performance of island models on an instance~(ft20) of Fisher and Thompson~\cite{fisher1963probabilistic}.

The TSP~\cite{hoffman2013traveling} assumes that there is a salesman travelling among $m$ cities from the starting city $s$ and returning to it. The constraint is that each city can be travelled only once. In this problem, the cost of travelling from city $i$ to city $j$ is denoted as $c_{ij}$, the best solution of this problem is to minimise the distance the salesman travels. We use an instance from TSPLIB\footnote{http://comopt.ifi.uni-heidelberg.de/software/TSPLIB95/} which is extended to 2000 cities.

The quadratic multiple knapsack problem considers that given a set of knapsacks of limited capacity and a set of objects, each object is associated with a weight, an individual profit and a pairwise profit with them. QMKP aims to maximize the profit being subject to the capacity constraint of each knapsack. The selected instance is from QKPInstance\footnote{http://cedric.cnam.fr/~soutif/QKP/QKP.html} and includes 20,000 objects and a density of 50\%.

\section{Experiment}
DIM-SP was coded in Java based on ECJLib\footnote{https://cs.gmu.edu/~eclab/projects/ecj/} and run on a PC with 3.2 GHz Intel(R) Core (TM) i7-3770 CPU and 8.00GB memory. In this study, the performance of DIM-SP was compared with the three other state-of-the-art island models with the same configuration of evolution operators and migration policy on JSSP, TSP and QMKP. In order to evaluate the performance of DIM-SP, we modified the datasets of these three problems. JSSP is a small scale problem with 200 jobs, TSP is a median scale problem with 2000 cities and QMKP is a large scale problem with 20,000 objects.

In the experiment, we set up the number of islands as ten for baseline island models as they are not dynamic. For the proposed island model, ten is the upper-limit of island
number. The island size is 200 and the maximum number of generations is 2000. All islands were assigned to the same evolution operators: crossover probability was 0.8, the mutation probability was 0.2 and migration frequency was 50 generations. In each migration, 5\% of randomly selected individuals migrate to the other connected islands evenly.

\begin{table*}[!t]
	\renewcommand{\arraystretch}{1.3}
	\caption{Experimental results of the proposed DIM-SP island model, Fully-connected island model, Star-shape island model and Ring island model on JSSP, TSP and QMKP problems.}
	\label{tab1}
	\centering
\begin{tabular}{|l||c|c|c||c|c|c||c|c|c|} \hline
	\multicolumn{1}{|l||}{\multirow{2}{*}{\textbf{Island Model}}}
	& \multicolumn{3}{|c||}{\textbf{JSSP}}
	& \multicolumn{3}{c||}{\textbf{TSP}}
	& \multicolumn{3}{c|}{\textbf{QMKP}} \\ \cline{2-10}
	\multicolumn{1}{|l||}{}
	& \textbf{Avg score}  & \textbf{Bst score}  & \textbf{Diversity}
	& \textbf{Avg score}  & \textbf{Bst score}  & \textbf{Diversity}
	& \textbf{Avg score}  & \textbf{Bst score}  & \textbf{Diversity}  \\ \hline
	DIM-SP
	& 1,216  & 1,180 & 0.291 & 13,962 & 13,225 & 0.347 & 28,937 & 29,286  & 0.416 \\
	Ring
	& 1,927  & 1,892 & 0.107 & 14,946 & 14,349 & 0.137 & 23,874 & 24,148  & 0.319 \\
	Star-shape
	& 2,479  & 2,136 & 0.107 & 15,278 & 14,436 & 0.136 & 21,943 & 22,822  & 0.289 \\
	Fully-connected
	& 2,877  & 2,623 & 0.107 & 15,917 & 15,112 & 0.136 & 20,932 & 21,327  & 0.271
	\\ \hline
\end{tabular}
\vspace{-0.2cm}
\end{table*}

The experimental results are shown in Fig.~\ref{fig6} where Fig.~\ref{fig6}a, Fig.~\ref{fig6}b and Fig.~\ref{fig6}c show the changes of average scores in different generations, which can be regarded as accuracy
estimation in traditional machine learning issues \cite{Wu:HDK12,Wu:TNNLS17}. Fig.~\ref{fig6}d, Fig.~\ref{fig6}e and Fig.~\ref{fig6}f show the changes of diversities in the benchmark problems.

In terms of scores, it can been seen that the proposed DIM-SP island model outperforms the other three island models. Generally, in the early generations, the scores of DIM-SP are worse than the other three models because it is initialized with just one island while the other models are initialized with ten islands. However, the performance of the proposed DIM-SP island model matches and overtakes the other models as evolution proceeds. This is because DIM-SP imposes high selection pressure on relatively mature sub-populations which enables it to find better solutions quickly. Specifically, JSSP, DIM-SP achieves lowest scores in both average (1,216) and best (1,180) scores followed by Ring model (average score1,927, best score 1,892), Star-shape model (average score 2,479, best score 2,136) and Ring model (average score 2,877, best score 2,623) because DIM can
effectively maintain high diversity compared to the other three. Similar patterns can also be found in TSP and QMKP. Another interesting finding is that Ring island model is generally better than Star-shape island model followed by Full-connected island model. This is because Ring island can maintain isolation among sub-populations for relatively longer periods of time compared to the other two models. This shows that it is important to create and maintain isolation as evolution proceeds to achieve good scores. DIM-SP can always generate a solution by clustering in epochs.

In terms of diversity shown in Fig.~\ref{fig6}d, Fig.~\ref{fig6}e and Fig.~\ref{fig6}f, DIM-SP can maintain diversity at a relatively high level compared to the other three island models. This is because the other island models tend to converge into the same sub-population while DIM-SP can maintain diversity by forcing sub-populations to be different via clustering. With DIM-SP, diversity is also stable as evolutionary proceeds. The reason is that we predefine the upper-limit of the number of islands. On the other hand, the topological features significantly contributes to diversity loss. The frequency of migration in Fully-connected island model is clearly larger than it is in Star-shape island model and Ring island model and sub-populations converge into the same one more quickly than the other two. This also reveals that one major weakness of island model is that they tend to converge into the same one because of migration. The DIM-SP model is proposed to overcome this to maintain different sub-populations effectively and efficiently.

Overall, with the same configurations of evolution operators of islands, the proposed DIM-SP island model outperform the other three island models in achieving high scores and maintaining diversities. The details are included in Tab.~\ref{tab1}.

\section{Conclusion}
In this paper, we propose a dynamic island model based on spectral clustering (DIM-SP) for overcoming the major drawbacks of traditional island models. The proposed model can force each island to maintain different sub-populations, control the number of islands dynamically and start with one sub-population. The experimental results confirms the proposed dynamic island model can achieve better performance than the other three baseline topological island models in both fitness score and diversity in three baseline optimization problems including job shop scheduler, travelling salesmen and quadratic multiple knapsack problems. We will further test the performance of the proposed island models in continuous domains with other population-based methods.
\bibliographystyle{IEEEtran}
\bibliography{myReferences}

\end{document}